\documentclass[a4paper, 10pt, conference]{ieeeconf}      

\IEEEoverridecommandlockouts                              
\usepackage{graphics} 
\usepackage{epsfig} 
\usepackage{amsmath} 
\usepackage{amssymb}  
\usepackage{multirow}
\usepackage{tabularx, booktabs}
\usepackage{pifont}
\usepackage{caption}
\usepackage{subcaption}
\newcommand{\mcheck}{\textcolor{dgreen}{\ding{51}}}%
\newcommand{\mcross}{\textcolor{dred}{\ding{55}}}%

\newcommand{\comment}[1]{}

\newcommand{\bI}{\mathbf{I}}

\newcommand{\bY}{\mathbf{Y}}

\newcommand{\bhY}{\hat{\mathbf{Y}}}

\newcommand{\mF}{\mathcal{F}}

\newcommand{\mL}{\mathcal{L}}
\newcommand{\mP}{\mathcal{P}}

\newcommand{\mbR}{\mathbb{R}} 

\usepackage{color}
\definecolor{purple}{rgb}{1,0,1}
\definecolor{orange}{cmyk}{0,0.5,0.8,0}

\definecolor{dgreen}{rgb}{0.0,0.6,0.0} 
\definecolor{dred}{rgb}{0.6,0.0,0.0}   

\usepackage{flushend}

\title{\LARGE \bf
Improving Map Re-localization with Deep `Movable' Objects Segmentation on 3D LiDAR Point Clouds
}

\author{Victor Vaquero$^{ 1 *}$  Kai Fischer$^{ 2 *}$ Francesc Moreno-Noguer$^{1}$ Alberto Sanfeliu$^{1}$ Stefan Milz$^{2}$
\thanks{*Equal Contribution from a research stay Victor Vaquero did in Valeo. }
\thanks{$^{1}$Institut de Rob\`otica i Inform\`atica Industrial, CSIC-UPC
Llorens i Artigas 4-6, 08028 Barcelona, Spain 
        {\tt\small vvaquero@iri.upc.edu}}%
\thanks{$^{2}$Valeo Schalter und Sensoren GmbH, Hummendorfer Str. 74, 96317 Kronach, Germany.
        {\tt\small kai.fischer@valeo.com}}%
}

\begin{document}

\maketitle
\thispagestyle{empty}
\pagestyle{empty}

\begin{abstract}
Localization and Mapping is an essential component to enable Autonomous Vehicles navigation, and requires an accuracy exceeding that of commercial GPS-based systems. 
Current odometry and mapping algorithms are able to provide this accurate information. 
However, the lack of robustness of these algorithms against dynamic obstacles and environmental changes, even for short time periods, forces the generation of new maps on every session without taking advantage of previously obtained ones. 
In this paper we propose the use of a deep learning architecture to segment \textit{movable} objects from 3D LiDAR point clouds in order to obtain longer-lasting 3D maps. This will in turn allow for better, faster and more accurate re-localization and trajectoy estimation on subsequent days. 
We show the effectiveness of our approach in a very dynamic and cluttered scenario, a supermarket parking lot. For that, we record several sequences on different days and compare localization errors with and without our \textit{movable} objects segmentation method. 
Results show that we are able to accurately re-locate over a filtered map, consistently reducing trajectory errors between an average of $35.1\%$ with respect to a non-filtered map version and of $47.9\%$ with respect to a standalone map created on the current session.

%

\end{abstract}


\section{INTRODUCTION}

Accurate localization is an essential component of autonomous vehicles and intelligent transportation systems, as it enables the accomplishment of further tasks such as path planning, safety navigation or obstacle avoidance.
Moreover, estimating a precise position in a map will also allow for obtaining further environmental information such as traffic state, accidents, or road closures/works, which would in turn facilitate the eventual completion of the predefined mission. 
The same idea holds in the opposite direction, in which a correctly located vehicle may augment map information with its current observations of a scene.

Nowadays, vehicle position can be easily obtained by different Global Navigation Satellite Systems (GNSS) such as GPS, Galileo, GLONASS, etc. Although these systems can provide good results, they have limited precision in urban scenarios with buildings and other elements that may block the satellite signals. Other accurate approaches like beacon-based methods exist, but require prior installation of external infrastructures and thus are not ready for general usage.

\begin{figure}[t]
\centering
 	\includegraphics[width=0.98    \columnwidth]{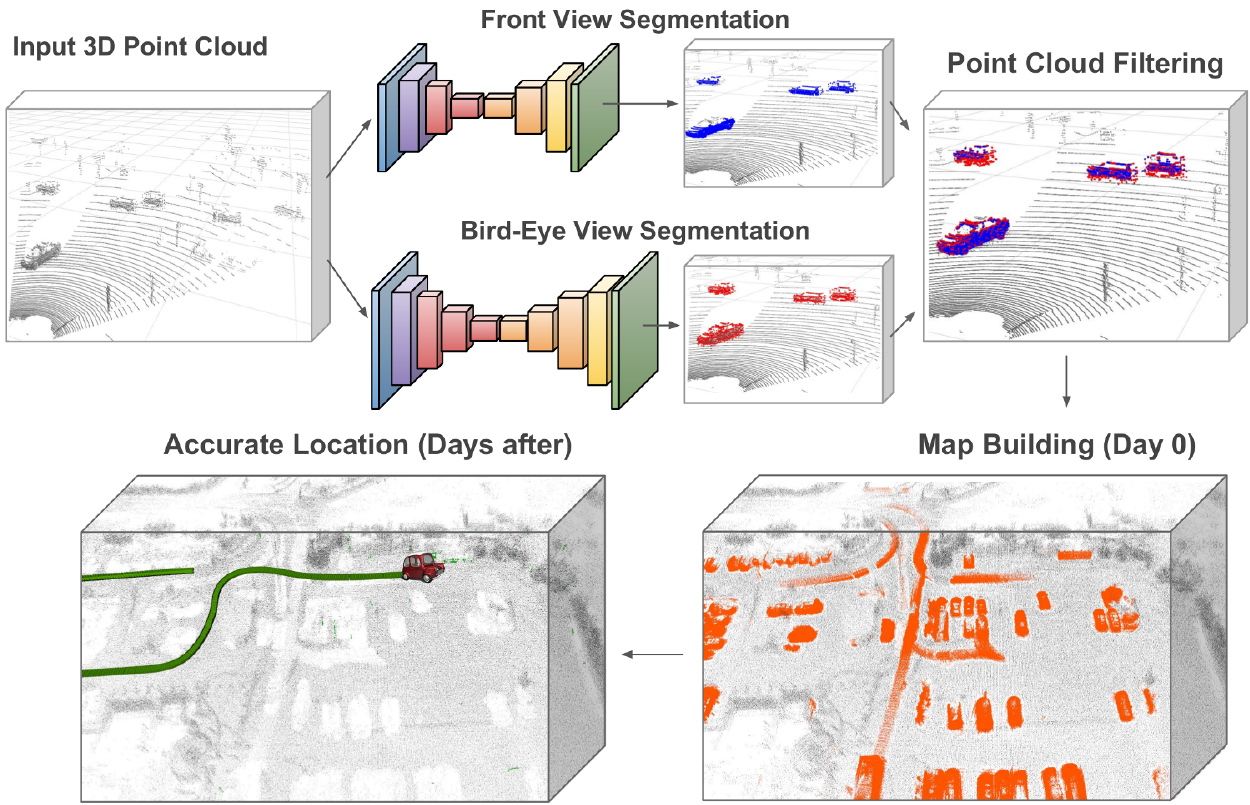}
 \caption{We propose to segment \textit{movable} objects from 3D LiDAR point clouds to build longer lasting maps that can assist on trajectory estimation for subsequent days. For that we employ two deep networks processing respectively a front and a bird's eye view projection of the LiDAR input frames. By retaining mostly static elements on the scene, we are able to accurately estimate our position and trajectory on subsequent days by additionally re-localizing on the map.}
 \vspace{-4mm}
 \label{fig:init}
\end{figure} 

For autonomous vehicles, it is preferable to include localization systems based on their own perceptive sensors, such as cameras or LiDAR. Although cameras can provide very rich information and colored textures, they are very sensitive to changes of illumination and lack of light. Contrary, LiDAR sensors provide robust and accurate 3D range measurements independently of the illumination conditions and even at night, which is why in this article we focus only on the use of these sensors.

Simultaneous Localization and Mapping (SLAM) has gained utmost attention within in-vehicle localization algorithms. However, in very dynamic and cluttered urban environments where vehicles and other elements are constantly moving or can potentially do, SLAM algorithms encounter difficulties on finding static and stable features that would allow to re-use the generated map on subsequent days.
Thus, in many applications where the goal is just to travel predefined routes in a known area, SLAM systems may introduce unnecessary and redundant computations creating a new map each time instead of just providing localization. 
Our motivation is therefore to find potential moving obstacles in the source point cloud data, excluding them from the mapping pipeline and allowing for longer standing representations while providing better accuracy.

With the advent of deep learning (DL), in-vehicle perception systems have strengthen their capacities on vital tasks for autonomous driving. 
Moreover, new developments are expanding the applicability of these techniques beyond optical cameras and postulating them also as powerful tools for working with 3D LiDAR data \cite{li2016vehicle, qi2017frustum, zhou2018voxelnet, vvaquero2018icra, beltran2018birdnet}. This fact also motivates us to leverage DL approaches for segmenting potential \textit{movable} objects on 3D LiDAR point~clouds.

To summarize, in this paper we propose the creation of longer lasting 3D LiDAR maps with state of the art localization and mapping algorithms by eliminating from the source the possible dynamic components of the scene.
%
For that, we take advantage of deep learning techniques and introduce the use of a two-stream deep convolutional architecture which, having respectively as input a front and a bird's eye view projections of the 3D LiDAR point cloud, segments the \textit{movable} objects from the scene. 
To demonstrate our approach, we build 3D maps of a very dynamic environment such as a parking lot with and without our segmentation applied and use it to assist for locating ourselves at different times and days, showing a consistent reduction of the average trajectory estimation error. 
Moreover, we show how our approach can also be employed for building a full map of an area covered in several days.
Our main contributions are: 
\begin{itemize}
\item{We present a deep convolutional dual-view architecture that having as input 3D LiDAR point clouds is able to segment potentially moving elements from a driving scenario, such as vehicles, cyclists or pedestrians.}

\item{We introduce a simple yet effective re-localization approach for odometry and mapping algorithms based on feature matching against a previously generated map, extending for longer time the life of those maps.}

\item{We show that by eliminating the dynamic elements in a scene, localization for subsequent days improve, demonstrating that most of the strong features extracted by current LiDAR odometry and mapping algorithms may lay on moving elements such as vehicles. } 

\item{We perform real experiments recording a parking lot scenario for several days and different trajectories, obtaining consistent quantitative results that support our approach. 
Additionally, we show the application of our method for building maps in a multi-agent manner or through different days.} 
\end{itemize}


\section{RELATED WORK}

In this section we review the state of the art of Deep Learning techniques and localization and mapping algorithms, both applied over 3D LiDAR point clouds. 

\vspace{1mm}
\noindent{\textbf{Deep learning applied on LiDAR point clouds.}}
Although Convolutional Neural Networks (CNNs) have been successfully applied on image-based tasks such as object classification, detection or semantic segmentation ~\cite{krizhevsky2012imagenet, ren2015faster, long2015fully} between others, its potential has not been yet extensively deployed to analyze 3D LiDAR point clouds. 
However, recent approaches are demonstrating the high capacities of deep neural networks to process LiDAR information on problems such as vehicle detection~\cite{li2016vehicle, qi2017frustum, zhou2018voxelnet, beltran2018birdnet} or motion segmentation~\cite{vvaquero2018icra}. 
In addition, the availability of large-scale real world datasets~\cite{geiger2012cvpr} as well as synthetic data \cite{Dosovitskiy17} are allowing the training of data driven deep models more easily.

Initial straightforward applications over 3D LiDAR data made use of 3D convolutions \cite{li20163d}, more efficient sparse convolutions \cite{engelcke2017vote3deep}, or just subdivided the input point cloud into voxels~\cite{zhou2018voxelnet}. 
%
Some other very recent approaches directly use the raw point cloud. In this way, PointNet~\cite{qi2017pointnet} applies a set of transformations and multi-layer perceptrons to generate global point cloud features which are then used for classification and segmentation tasks. PointNet++~\cite{qi2017pointnet++} proposes to recursively apply PointNet on nested areas of the input point cloud, learning additionally local features with increasing contextual scales. 
Frustrum-PointNet~\cite{qi2017frustum}, explores larger areas and extracts 3D frustums from bounding boxes given by a 2D CNN detector over RGB images. However, all these methods required of high computational power and include ad-hoc steps that can greatly affect its performance and stability.

Nowadays the most embraced procedure is to project the 3D LiDAR point cloud to create 2D representations from which to apply standard image-based 2D convolutions. 
In this way,~\cite{li2016vehicle} uses a front view projection encoding the range distance and height of each 3D point and train a deep network to extract vehicle 3D bounding boxes. Similarly, \cite{vaquero2017deconvolutional} creates a front view projection of the polar LiDAR coordinates along with the reflectivity of each point, to segment vehicles by predicting the \textit{vehicleness} confidence of each point.
On the other hand, BirdNet\cite{beltran2018birdnet}, TopNet~\cite{wirges2018object} or RT3D~\cite{zeng2018rt3d} make use of a bird's eye view projection of the point cloud, encoding different features on each cell. 
%

In this paper we devise a dual-branch LiDAR convolutional architecture for filtering all the potentially \textit{movable} objects in the scene. 
One branch processes a front view of the 3D LiDAR input whereas the other a bird's eye view projection, and each of them predicts the probability of the 3D projected points (\textit{pixels}) to belong to a \textit{movable}  class (e.g. car, bicycle, pedestrian, etc.) or otherwise \textit{non-movable}.

\vspace{1mm}
\noindent{\textbf{Map building and localization.}}
%
Localization and Mapping with LiDAR is a very active research topic in the robotics and automotive community. 
Early approaches used 2D laser data and ICP methods to correct the ego-motion distortion. 
However, when employing more complex 3D LiDARs with further amounts of information more sophisticated approaches need to be considered, like including other sensor's information such as IMU, wheel encoders or GPS/INS using for example extended Kalman filters~\cite{moosmann2011velodyne, wang2013lidar}. 

Some early approaches take motivation from visual SLAM methods~\cite{durrant2006simultaneous} and create intensity images using the laser returns from which to extract and match distinctive features between frames to infer the ego motion~\cite{dong2014lighting, anderson2013ransac}. 
However, in these methods based in matching visual or geometric features the localization and trajectory is usually recovered by a batch optimization post-process, which make them unsuitable for real time localization. 

\begin{figure*}[t]
\centering
 	\includegraphics[height=4cm, width=0.95\textwidth]{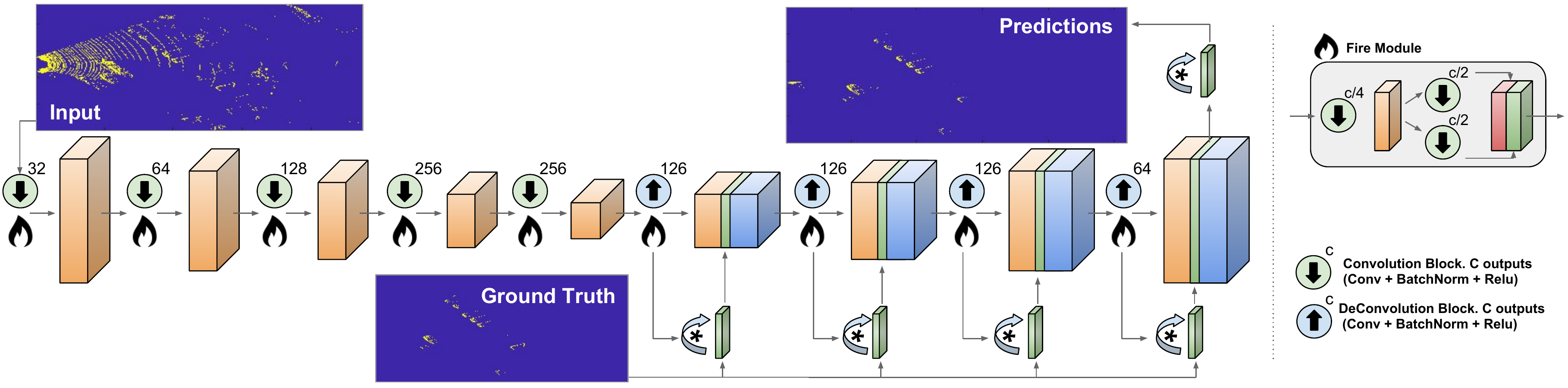}
 \caption{We model $\mF_{BE}$ as a refined encoder-decoder architecture. As the inputs are quite big ($500 \times 600$) we apply five contractive and five expansive levels. To get richer features at each level, we insert customized fire modules that capture local and context information from the previous feature maps. These modules first reduce the number of feature channels and apply two parallel sets of convolutional filters on them to finally concatenate the results, obtaining local and context aware features. Intermediate losses are computed in this network, merging predictions at different resolutions.}
 \vspace{-5mm}
 \label{fig:be_net}
\end{figure*}

The introduction of LOAM algorithm~\cite{zhang2014loam} supposed a great advance in terms of accuracy --achieving top position on the KITTI Odometry benchmark-- as well as real-time performance. It proposes to divide the complex SLAM problem in two algorithms, where one estimates odometry at a high frequency but low fidelity whereas the other runs at lower frequencies for fine matching and registration of the point cloud. 
From this approach, incremental improvements have been done. LeGO-LOAM~\cite{shan2018lego}, proposed a lightweight, real-time pose estimation method that has into account the presence of a ground plane in its segmentation and optimization steps to obtain distinctive planar and edge features for the next optimization method. We base on this approach to build our maps as it provides a real-time and accurate representation as well as is more robust using real raw LiDAR data.

Other contemporary approaches propose to infer the vehicle position by matching segments from the point cloud belonging to partial or full objects as well as sections of larger structures~\cite{segmatch2017}. In this way they obtain a good balance between local descriptors, which may suffer from ambiguity without context information, and global features, which are viewpoint dependent. The method extract several features from the segmented clusters and try to match those segments over the ones existing in a map. 
Further incremental approaches employ a data driven feature encoder to extract compact and discriminative features from the segments~\cite{segmap2018}, which can be used for creating a compact map representation due to its high reconstruction capacities and for accurate location over an existing map, as long as it does not contain much movable elements.

In this way, the ability to create longer standing maps overcoming challenging issues such as map scalability and updatability or management of dynamic elements, is still a pending task for SLAM algorithms \cite{cadena2016past, bresson2017simultaneous}
In the presented work, we directly segment the input information with a deep learning pipeline prior to build any map. Thus, we avoid from the beginning the inclusion of possible outliers and known-dynamic elements, so therefore do not give chances to further extract any feature from them. By doing so, we demonstrate in Section~\ref{sec:exps} that more accurate localization is possible on a constantly changing environment over a map created days before.






\section{APPROACH}

The main objective of this paper is to demonstrate how, by pre-filtering possibly moving objects from the scene, we are able to accurately locate during longer periods of time in standard 3D maps such as the ones generated with common SLAM or LiDAR odometry and mapping (LOAM) algorithms. 
Our approach has two main modules, which are detailed in this section: 
A) deep-learning based segmentation of movable objects in 3D LiDAR point clouds; 
B) accurate re-localization at different days over a map built from a cluttered dynamic scenario using standard LeGO-LOAM algorithm. This can also be employed to build a full map over several days or in a multi agent way.


\subsection{Movable Objects Segmentation}
For this task, our only input is a 3D LiDAR point cloud $\mP = \{q_1,\cdots,q_Q\}$, where each point $q_k \in \mbR^4$ is represented by its Euclidean coordinates and the returned reflectivity. 
Our objective is to classify each of these 3D points as belonging to a $\{$\textit{movable}, or \textit{non-movable}$\}$ class, where we consider as \textit{movable} all the Kitti~\cite{geiger2012cvpr} annotated classes, i.e. `Vehicle', `Van', `Truck',`Cyclist',`Pedestrian', `Tram' and `Misc'.

We formulate the task as a binary semantic segmentation one, in which we perform a per-point classification prediction. To solve it, we take advantage of the recent success of deep convolutional neural networks and propose to model two segmentation functions $\mF_{FR}$ and $\mF_{BE}$  (see Fig.~\ref{fig:be_net}), each one respectively committed over a front view $\bI_{FR}$ and a bird's-eye view $\bI_{BE}$ 2D projections of the 3D LiDAR data. Therefore we want to learn the $\mF$ mappings such that:
\begin{equation}
\begin{split}
	\mF_{FR} : (\bI_{FR}, \bY_{FR}) \rightarrow \bhY_{FR} \\
	\mF_{BE} : (\bI_{BE}, \bY_{BE}) \rightarrow \bhY_{BE}
	\end{split}
\end{equation}
where
$\bY_{FR}$ and $\bY_{BE}$ are two ground truth masks that indicate whether or not each 3D projected point belongs to a \textit{movable} object and $\bhY_{FR}$ and $\bhY_{BE}$ will be the predicted probability maps on each projection plane.
Next, we present the main components of the stated deep segmentation approach, such as the projections and ground truth created from the 3D point cloud, the convolutional architectures involved and the training process to model our objective functions.

\vspace{1mm}
\noindent{\textbf{Front View Projection.}} The input Front view, $\bI_{FR}$, is obtained similarly than in~\cite{vaquero2017deconvolutional}. We arrange the 3D point cloud $\mP$ according to the Velodyne HDL-64 geometry into a 2D array such that $\bI_{FR} \in \mbR^{H \times W \times C}$. 
The Euclidean points $(x,y,z)$ are transformed to spherical coordinates, where the elevation angle $\theta$ represent the $H=64$ horizontal lasers of the Velodyne sensor. 
We filter the point cloud in the corresponding camera Field of View (FoV) that contains the KITTI annotations ($\phi \in [-40.5,40.5]$) and discretize it according to the sensor manufacturer using an azimuth step of $\Delta\phi=0.18$ degrees, which map to a width of $W=448$ pixels.
In the third dimension of our $\bI_{FR}$ front view map we store the corresponding range values $\rho$, and reflectivity $r$, so therefore obtaining $C=2$ channels.

\vspace{1mm}
\noindent{\textbf{Bird's Eye View Projection.}} 
The bird's eye (zenithal) view  $\bI_{BE}$ is obtained over an area of $60 \times 50$ meters in front of the LiDAR sensor after carefully observing that roughly $95\%$ of the annotated vehicles in Kitti are within these margins. 
Inspired by~\cite{beltran2018birdnet, wirges2018object}, we generate a 2D grid with a resolution of $0.1$ meters and project the cropped point cloud on it. We consequently obtain a bird's eye view $\bI_{BE} \in \mbR^{H' \times W' \times C'}$, where $H'=600$, $W'=500$, and $C' = 6$, accounting for six different features:
1) a binary occupancy term with zero value if no points are projected in the cell and one otherwise; 
2) an absolute occupancy term, counting the total number of points in the cell; 
3) the mean reflectivity value of the points on the cell; and 4, 5, and 6) the mean, minimum and maximum height values of the points projected on the cell. 

\vspace{1mm}
\noindent{\textbf{Ground Truth Generation.}} 
The \textit{movable} elements ground truth for the front $\bY_{FR}$ and bird's eye $\bY_{BE}$ projections is generated by using the 3D-oriented bounding boxes from the KITTI Tracking dataset. We transform these bounding boxes from the camera to the LiDAR frame and label the 3D points that fall inside each box of all the annotated \textit{movable} classes.

\vspace{1mm}
\noindent{\textbf{Network Architectures.}}
We propose for both projection domains contractive-expansive architectures which allows for a good embedding of features. 
Additionally, we include skip connections and concatenate feature maps between contractive and expansive parts to build stronger features that will help the learning process by back-propagating purer gradients from the upper parts to the lower layers.

\subsubsection{Front view Architecture}
We employ for the front view segmentation task the deconvolutional architecture proposed in~\cite{vaquero2017deconvolutional}. 
As key design, it imposes a stride ratio of 1:2 in the first convolutional layer to obtain more tractable intermediate feature maps, reducing the input size imbalance of $\bI_{FR}$ from [64 vs 448] to [64 vs 224]. From there, it performs two more resolution decreases in the contractive part of the network, which are later recovered on the expansive sector.
Additionally, in this network we also carefully design the initial filter sizes according to the observed shape of the most predominant \textit{movable} object in this view (vehicles), so imposing a filter of 7x15 with a big initial receptive field.

\subsubsection{Bird's Eye View Architecture}
Within this view we encounter a new challenge which is that, averaging within the training set with the chosen grid resolution of $0.1$ meters, movable objects occupy less than the $8\%$ of the grid-cells per frame. 
Segmenting these small areas is still a challenging problem for deep neural networks, and force us to design the specialized architecture shown in Fig.~\ref{fig:be_net} to manage the trade-off between accuracy, number of parameters and speed. 

To keep the number of network parameters small while still providing high accuracy, we employ in our architecture a customized version of the well-established convolutional `fire modules'~\cite{iandola2016squeezenet, Bwu2018squeezeSeg, wu2018squeezesegv2}, which can be seen in the top area of Fig.~\ref{fig:be_net}. 
Within these modules, we initially use a convolution layer to reduce the number of feature maps and later on we parallelly apply two new convolutional layers with different filter sizes. 
Results are finally concatenated to obtain robust features with local and large context-aware information, while using a low number of parameters. 
In our architecture we include these `fire modules' after each change of resolution of the feature maps. 

\vspace{1mm}
\noindent{\textbf{Training the Networks.}}
We train both networks to segment the front and bird's eye views in a supervised manner using a class weighted cross entropy loss function~\cite{vaquero2017deconvolutional}, defined as:
\begin{equation}
\small
\mL^{\textit{WCE}}(\bI^n, \bY^n) = -\sum_{h,w,l}^{H,W,L} \omega(Y^n_{h,w}) \textit{Id}_{[Y^n_{h,w}]} \text{log}(\mF(I^n,Y^n)_{h,w,l}), 
\label{eq:2}
\end{equation}
where $\bI^n$ is the $n$-th training projection sample and $\bY^n$ is the corresponding ground truth map. We compute a class imbalance weighting function $\omega$ as the inverse ratio between the vehicle and background classes from the training set samples. $\textit{Id}$ 
is an index function that selects the predicted probability associated to the expected ground truth class.

In order to guide the network to a correct result faster, we implement a multi-scale solution for the segmentation problem by introducing intermediate predictions and losses at different resolutions, which insert valuable gradients at middle levels.  
Hence, we compute the final loss $\mL$ independently for the front and bird's eye networks as:
\begin{equation}%
\label{eq:CaF}
	\mL(\bI^n, \bY^n) =
	\sum_{m=1}^{M} \lambda_{m} \mL^{\textit{WCE}}(\bI^n_m, {\bY^n_m})
\end{equation}
where $\lambda_{m}$ are regularization weights for each resolution loss and $m$ is the resolution step at which the loss function is computed. We compute $3$ partial losses for the front view projection network, and  $5$ for the bird's eye view one.

We train our networks using the KITTI Tracking dataset from which we obtained the ground truth of movable objects. 
To preserve the geometry properties of the driving scene we augment the dataset with horizontal flips in the front view and vertical ones in the bird's eye view with a $50\%$ chance. 
For the training procedure, we initialize the architectures with He's method~\cite{he2015delving} and use Adam optimization with parameters $\beta_1 = 0.9$ and $\beta_2 = 0.999$. 
We train each network independently on a single Nvidia 1080Ti GPU using a batch size of $10$ during $400,000$ iterations. We start with a learning rate of $10^{-3}$, which is halved every $50,000$ iterations after the first $150,000$. Attending to the ratio between classes on each domain, we set the regularizator $\omega$ to $25$ in the front view and to $1000$ in the bird's eye view network. The multi-resolution loss regularizers $\lambda_r$ are set to $1$, assigning equal importance to each resolution.

\vspace{1mm}
\noindent{\textbf{Filtering Movable Objects.}}
To filter out the points that belong to \textit{movable} objects in the 3D input LiDAR point cloud, we first obtain the segmentation prediction from the two deep models and fuse the obtained predictions back in the 3D Euclidean space. 
Next we cluster points from both predictions and validate them according to a minimum size of at least 50 points per cluster. We also discard clusters according to its mean probability predicted, where we weight the contribution of the bird's eye samples to be twice as the front view ones ($0.2$ vs $0.1$ respectively) as this last have shown noisier results, and set a threshold of $0.13$ to approve resulting clusters. 
Once we have the final clusters of the predictions, we filter the original input LiDAR data by eliminating all points in a radio of 10 cms from a predicted one, so that restricting the effect of possible projection errors.

\subsection{Vehicle Localization}

To locate inside an already built map our system uses just Velodyne LIDAR data and, if available, information from low-cost car GPS. 
Although the latter is very imprecise and has a very low refresh rate, it is a standard equipment nowadays in most vehicle, so we can use it to estimate a coarse initial position which is afterwards refined by using our re-localization algorithm. 
Next, we first present the creation of the ground truth map at day zero against which we aim to locate on the subsequent days, and then we detail our localization approach, which consist on obtaining an coarse initial guess followed by the final accurate localization. 



\vspace{1mm}
\noindent{\textbf{Ground Truth Map Building.}
\label{map_building}
For building the ground truth map from which to validate our approach, we employ DGPS-synchronized LiDAR scans that are feed to the state of the art LeGO-LOAM algorithm in charge of composing the final map representation. 
LeGO-LOAM also extracts edge and surface features from the generated map by analyzing the local surface properties of certain areas in the point cloud. Edge features are extracted from rough local regions, whereas surface ones are collected from smooth surfaces.

In order to obtain more distinctive features from the map, the LeGO-LOAM algorithm does not account for features within a minimum distance from others considered as strong. This fact have a big counterpart; as movable objects like vehicles have a very prominent surface structure, there are therefore more likely to be chosen as strong features over other relevant static features of the environment. 
In this way, by pre-filtering the point cloud we enforce the selection of distinctive strong features just from static elements instead of from movable objects, thus allowing better inter-day re-localization. A comparison of the extracted features from the full and the filtered point cloud respectively can be observed in Fig. \ref{fig:car_features}
Additionally, we also remove the ground-floor features determined by LeGO-LOAM to obtain a more compact and distinctive representation of scene.

Finally, our resulting ground truth map (GT-Map) consists of LeGO-LOAM features extracted from each filtered frame along with the own frame transformation from the DGPS. 



\begin{figure}
\centering
     \begin{subfigure}[b]{0.48\textwidth}
         \centering
         \includegraphics[width=\textwidth]{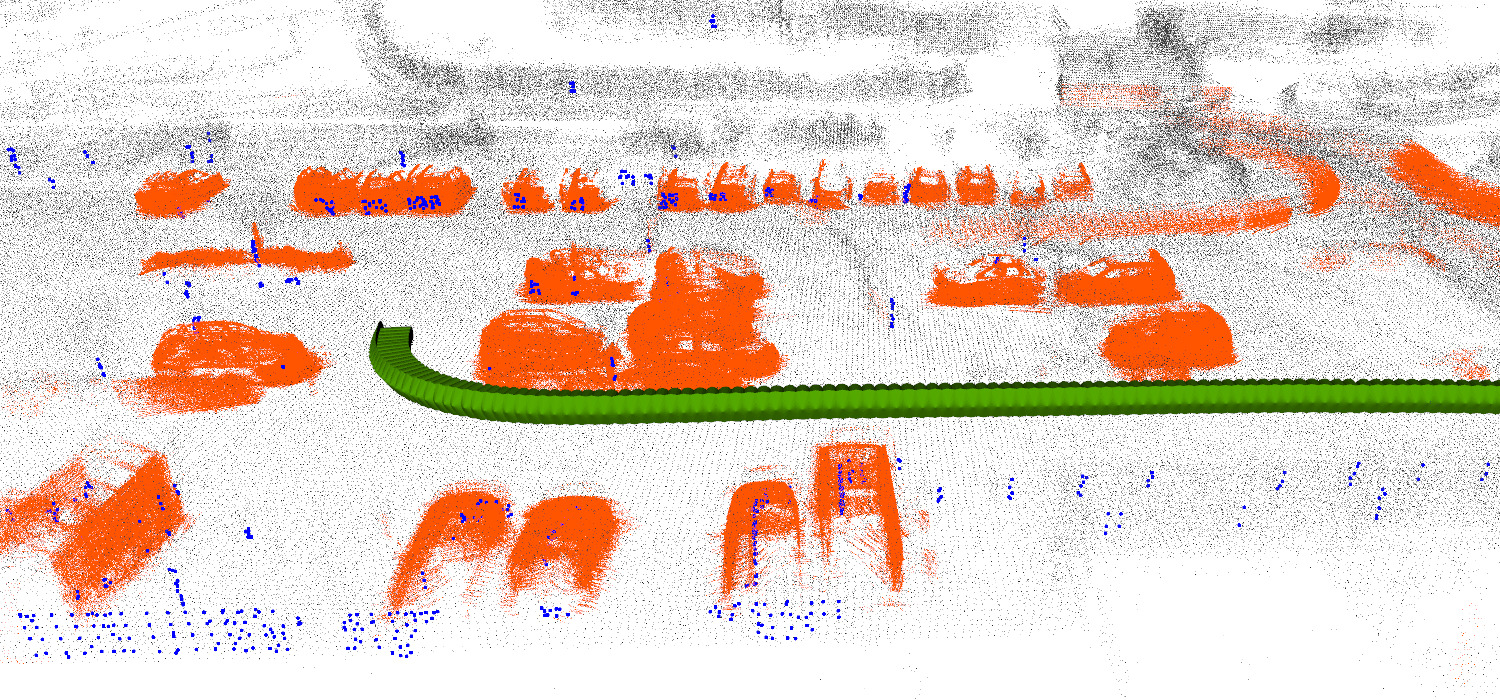}
         \vspace{-7mm}
         \caption{Features extracted from full point cloud}
         \label{fig:car_features_bad}
     \end{subfigure}%
     \vspace{1mm}
     \hfill
     \begin{subfigure}[b]{0.48\textwidth}
         \centering
         \includegraphics[width=\textwidth]{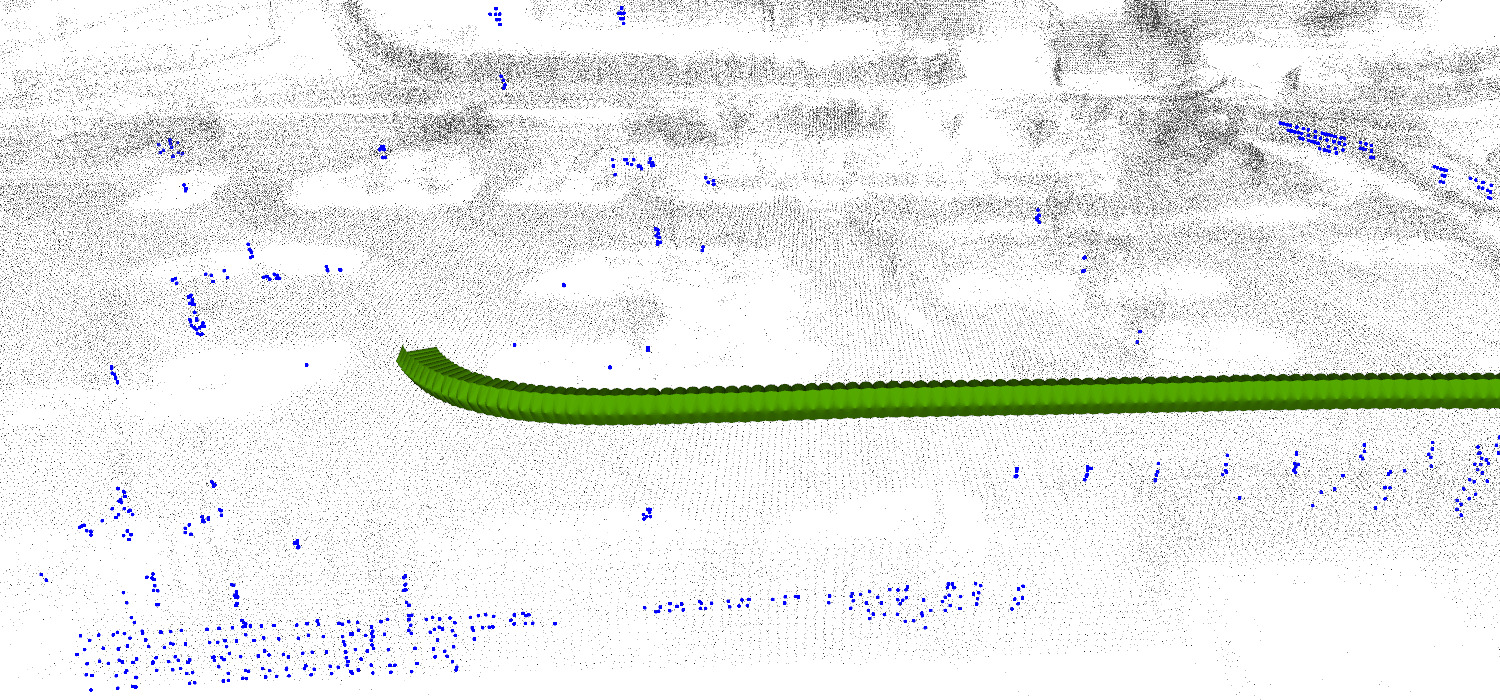}
         \vspace{-5mm}
         \caption{Features extracted from filtered point cloud}
         \label{fig:car_features_good}
     \end{subfigure}
        \caption{Comparison of the extracted features (blue) from the full point cloud and the one with removed movable objects respectively. By providing an unfiltered point cloud the feature extraction mechanism selects a vast amount of points from dynamic objects which can be compensated by applying our proposed movable object detection algorithm.}
        \vspace{-5mm}
        \label{fig:car_features}
\end{figure}

\vspace{1mm}
\noindent{\textbf{Initial Pose Estimation.}}
To initially accelerate the current localization process over a previously existing map, we can use any rough prior information. In our experimentation, we use a commercial GPS available nowadays as standard on-vehicle equipment. Notice that we use it only to speed up the initial localization, but it is not strictly necessary. 

As GPS solely provides coarse information about the position, we additionally need to estimate the initial orientation. 
In order to do that, we extract a subset of our ground truth map around the GPS coordinates and perform feature matching against our current observed frame using ICP. Feature points used for the current frame are extracted and selected in the same way as described in the map building process. 
To optimize the ICP process and get an initial orientation estimation, we firstly perform a coarse matching prediction by applying ICP over different rotations of the current frame features on steps of 45 degrees. For each rotation, we get a fitting score, which describes the remaining sum of squared differences from the feature points of the current frame to their corresponding nearest neighbours in the ground truth map. Our initial orientation guess is selected as the one with a fitting score lower than $0.4$.

Finally, the transformation to our initial pose estimation can be expressed as $T_{init} = T_{GPS}  \cdot T_{ICP} \cdot T_{Rot}$, where $T_{GPS}$ would be initial rough position estimate given by the commercial GPS, $T_{Rot}$ is the best fitting initial rotation found for the ICP, and $T_{ICP}$ refers to the final refined transformation obtained by the ICP algorithm that optimizes the feature matching best. Notice here that $T_{GPS}$ and $T_{Rot}$ are just used to speed up the matching process, and that any other prior coarse pose estimation could be employed here.

\vspace{1mm}
\noindent{\textbf{Continuous Re-Localization}}
Once the initial pose estimation is performed, we keep employing LeGO-LOAM to continuously calculate further transformations based on our segmented LiDAR scans. 
At the same time, we perform re-localization, trying to match our current position against the pre-existing generated map (GT-Map). 
For this continuous re-localization steps, we follow a similar process than above and employ the extracted features from the current scan with ICP to correct possible drifts caused by the current trajectory obtained with the LeGO-LOAM algorithm. 

In comparison to the initial pose estimation, these re-localization transformations are simpler to calculate, as they only depend on the current estimated position and the correction given by the ICP over the GT-Map. Therefore, $T_{reloc} = T_{ICP} \cdot T_{c }$, where $T_c$ stands for the current pose estimated. The threshold for the ICP fitting score in the course of this re-localization step is lowered to $0.3$ in order to estimate the transformation more robustly.


\section{EXPERIMENTS}
\label{sec:exps}


To show the effectiveness of the proposed approach, we have recorded 7 different sequences of a cluttered and dynamic urban environment, i.e. a supermarket parking lot, on different days and at diverse hours. 
In this section we first detail the data acquisition process for the experiments. Next, we show our re-localization capacities in this highly unsteady scenario using as GT-Map a recorded sequence from a different day that would be useless if not for our approach, as it would not last more than the session for which it was created. We show additional applications of our approach for building a map through different days, which can also extrapolates to multi-agent map building tasks.


\begin{figure}[t]
\centering
 	\includegraphics[width=0.97\columnwidth]{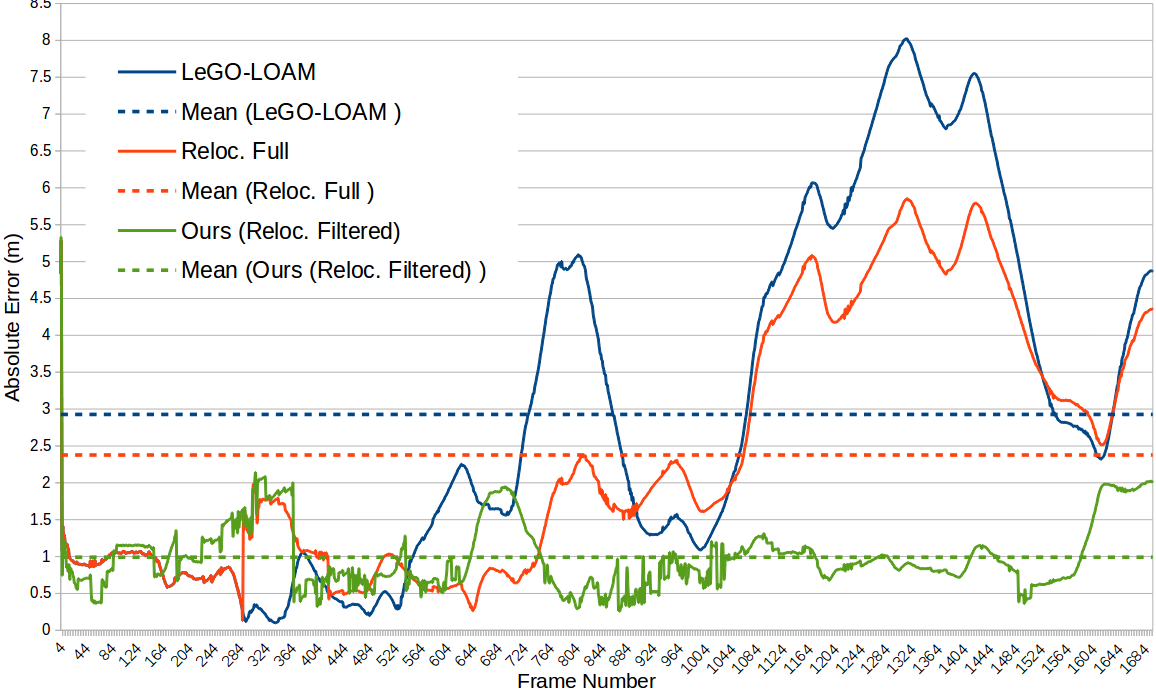}
 \caption{Progression of the absolute error of re-locating Sequence 3 in Map 1 for the three regarded methods in comparison to the ground truth data including each corresponding Mean Absolute Errors (MAE) as dotted line.}
 \vspace{-5mm}
 \label{fig:error_plot}
\end{figure} 

\begin{figure}
\centering
     \begin{subfigure}[b]{0.49\columnwidth}
         \centering
         \includegraphics[width=\textwidth]{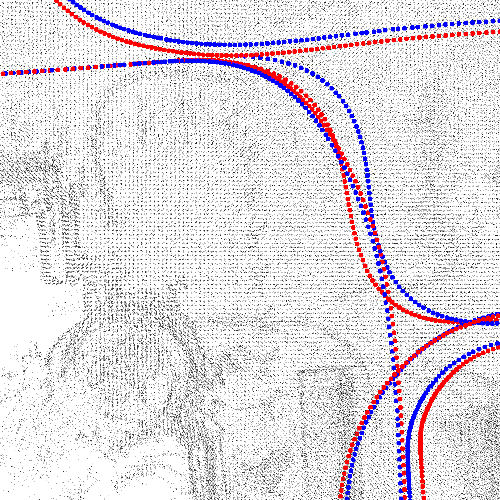}
         \caption{Drift using Reloc. Full}
         \label{fig:map_ext_bad}
     \end{subfigure}%
     \hfill
     \begin{subfigure}[b]{0.49\columnwidth}
         \centering
         \includegraphics[width=\textwidth]{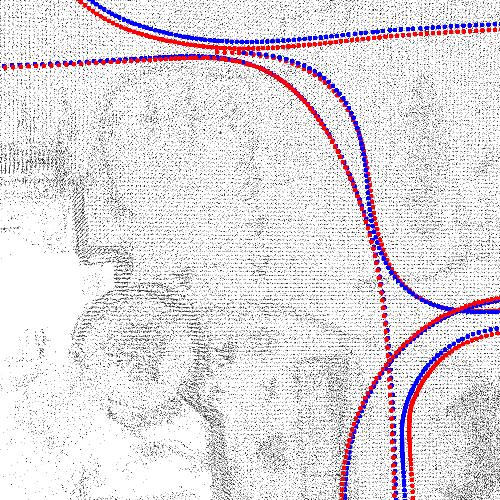}
         \caption{Better with Reloc. Filtered}
         \label{fig:map_ext_good}
     \end{subfigure}
        \caption{Qualitative results comparing the performances of `Reloc. Full' and `Reloc. Filtered' on the Map Extension experiment. In red we display the estimated trajectories along the days and in blue the ground truth. For `Reloc. Full' the resulting map shows blurry and doubled map contents caused by drift from the ground truth due to incorrect feature matching which are clearly compensated by our approach.}
        \vspace{-5mm}
        \label{fig:map_ext}
 \label{fig:map_extension}
\end{figure}

\vspace{-1mm}
\subsection{Data acquisition}
\vspace{-1mm}
The data used for evaluating our experiments was captured and recorded with a Test car by Valeo which is equipped with multiple sensors. The relevant hardware concerning our experiments is a Velodyne LiDAR HDL-64E S3, a differential GPS by IMAR and the serial production car GPS. 
The 7 sequences were recorded on a parking lot of a local shopping mall in Kronach (Germany) at different days and times to ensure diverse constellation of the parking cars. Sequence 1 was recorded early in the morning aiming to obtain an almost empty parking lot. Sequence 2 and 3 were recorded on another day with the area being slightly crowded. Sequence 4 to 7 were recorded on different hours of another day with a very crowded parking lot. Our dataset therefore consists of 7 sequences recorded at three different days with diverse constitution of the parking lot.

\vspace{-1mm}
\subsection{Re-localization in dynamic environments}
To validate our method we built two GT-Maps as described in \ref{map_building} using sequences 1 (GT-Map 1) and 3 (GT-Map 2). 
In this regard sequences 2 to 7 are used for re-localization over GT-Map 1 and sequences 1 and 4 to 7 over GT-Map 2, ensuring inter-day experiments with different environmental constitution. 
Moreover, for each sequence we build two GT-Map variants, a Full one including all objects, and a Filtered one with removed \textit{movable} elements.  

In our experiments, we apply three different methods: 
\begin{itemize}
    \item{\textbf{LeGO-Loam:} in which subsequent frames are processed in an unfiltered map solely based on the pose estimation calculated by LeGO-LOAM.}
    
    \item{\textbf{Reloc. Full:} in which we perform in parallel pose estimation by LeGO-LOAM and re-localization over an unfiltered map using the full unfiltered current frames.}
    
    \item{\textbf{Ours (Reloc. Filtered):} in which we perform in parallel pose estimation by LeGO-LOAM and re-localization over a pre-filtered map using the filtered current frames.}
\end{itemize}

\begin{table*}[th!]
 	\caption{Re-Localization results comparing LeGO-LOAM, Reloc. Full and Ours (Reloc. Filtered)}
 	\label{tab:reloc_results}
 	\centering
 	\begin{tabular}{c c  c c c  c c c  c c c c}
 		\toprule
 		 \multirow{2}{*}{GT-Map} & \multirow{2}{*}{Curr. Seq} & \multicolumn{2}{c}{LeGO-LOAM}  & \multicolumn{3}{c}{Reloc. Full} & \multicolumn{3}{c}{Ours (Reloc. Filtered)} &\multicolumn{2}{c}{MAE Improvements (\%)} \\  	
 		\cmidrule(lr){3-4} \cmidrule(lr){5-7} \cmidrule(lr){8-12}
 		 & & $1st_{r}$ & MAE (m) & $\#_{r}$ & $1st_{r}$ & MAE (m) & $\#_{r}$ & $1st_{r}$ & MAE (m) & To Reloc.Full & To LeGO-LOAM\\ 
		
		\midrule    \multirow{6}{*}{1} 
		& 	2	& 3 	& 1.73 & 7	& 3 	& 1.29 & 288	& 8 	& 0.84 	& 34.42 \% 	& 51.07 \%	\\ 
		& 	3	& 2 	& 2.93 & 20	& 2 	& 2.38 & 257	& 2 	& 0.99 	& 58.30 \%	& 66.13 \% \\ 
		& 	4	& 197 	& 2.09 & 0	& 275 	& 6.38 & 108	& 8 	& 1.08 	& 83.05 \%	& 48.23 \% \\ 
		& 	5	& 273 	& 3.65 & 0	& 274 	& 3.74 & 78		& 233 	& 1.98	& 47.03 \%	& 45.76 \% \\ 
		& 	6	& 567 	& 5.94 & 0	& 279 	& 4.14 & 5		& 281 	& 3.91	& 5.56 \%	& 34.25 \% \\ 
		& 	7	& 33 	& 3.59 & 0	& 27 	& 4.05 & 49		& 26 	& 1.01 & 75.05 \%	& 71.84 \% \\ 
 		\midrule    \multirow{5}{*}{2} 
		& 	1	& 2 	& 2.80 & 503& 2 	& 0.74 & 548 	& 2		& 0.70  & 5.02 \%	& 75.03 \% \\ 
		&	4	& 67 	& 2.26 & 14	& 66 	& 1.88 & 101 	& 56	& 1.73  & 7.91 \%  & 23.40 \% \\ 
		&	5	& 273 	& 2.78 & 54	& 289 	& 2.73 & 112 	& 269	& 2.12  & 22.32 \% & 24.32 \% \\ 
		& 	6	& 168 	& 4.73 & 9	& 168 	& 4.61 & 26		& 165 	& 2.39 	& 48.29 \% & 49.51 \% \\ 
		& 	7	& 1 	& 1.93 & 53	& 1 	& 1.20 & 135	& 1 	& 1.21 	& -0.79 \%  & 37.37 \% \\ 
 		 
 		\bottomrule
 	\end{tabular}    
 \end{table*}

For each method, performances in consideration of localization in a dynamic pre-build map are validated based on three different metrics:
\begin{itemize}
    \item{\textbf{First relocation} ($1st_{r}$): defined as the frame number of the initial pose estimation.}
    
    \item{\textbf{Number of relocations} ($\#_{r}$): number of frames at which the particular algorithm was able to relocate.}
    
    \item{\textbf{Mean Absolute Error} (MAE): in meters, the averaged absolute error of the estimated positions over the whole sequence compared to ground truth poses of the DGPS.}
\end{itemize}

Table \ref{tab:reloc_results} shows the re-localization performances of the considered methods and sequences applied on the respective maps. Additionally, we show in Fig.~\ref{fig:error_plot} the absolute error obtained with the three algorithms for re-localization along sequence 3 using the GT-Map from sequence 1 (almost empty parking map) in comparison to the ground truth data.


At the lights of the results we can observe that filtering \textit{movable} objects from the point clouds greatly improves the performance of re-localization in dynamic environments. Compared to the unfiltered re-localization (Reloc. Full, in Table~\ref{tab:reloc_results}) we reduced the MAE value a 16.5\% using GT-Map 1 and up to 50.57\% using GT-Map 2 averaging over the respective sequences. 
Additionally our approach consistently scores highest in number of localized frames and mostly gets faster initial localization compared to the unfiltered approaches. 

Since GPS data is used solely for localization until the initial pose estimation is obtained, there are bigger MAE values when this initial localization takes place late. 
The consequences of this effect can be observed in Table~\ref{tab:reloc_results} when re-locating at sequences 5 or 6 over both maps. Observe how over these sequences the first relocation occurs rather late, and therefore the MAE error is higher. 
This impact is mainly caused by partially nonexistent sequence and map overlap so therefore re-localization cannot be applied. 
The contrary effect can also be noticed in sequence 7 applied over GT-Map 2, where unfiltered re-localization ranks higher in MAE than our approach. In this occasion, similar constellation of vehicles were present in the parking, so for the two baseline algorithms the first re-localization is performed fast. 

Another factor affecting the shown results are partially erroneous re-localizations essentially happening in curves where a slight deviation in orientation estimation has a huge impact on the subsequent trajectory calculation. \mbox{Assuming} no further re-localizations occur after an erroneous re-localization there will be a huge drift in the ensuing poses which is reflected in the results of the unfiltered re-localization whose MAE is partially exceeding the ones of the standard LOAM approach. Comparing to the results of our approach this perturbation can be mostly eliminated, since in the filtered environment more distinctive, static features are selected to guarantee more robust pose estimations.

\subsection{Multi day map extension}
 Apart from experiments on re-localization in dynamic environments we can prove the adaptability of the proposed algorithm to the application of a mapping process during several days. Here we are able to show that filtering movable objects from the processed data drastically improves the ability to build correspondences between maps from different days and consequently, the quality of the final map. 
 In our experiments we choose sections of sequences from three different days which are partially overlapping to compose a final mapping of the entire parking lot.

 Starting with a segment of sequence 1 we are building the map solely using the LeGO-LOAM algorithm with loop closure to accomplish a detailed mapping result. 
 Next we are processing a section of sequence 3 and do re-localization in the previous built map based on the extracted features. In contrast to the previous experiments where we were doing a re-localization pose estimation separately for every frame here we are using the found correspondences to the previous map to do a graph optimization based on the inbuilt loop closure method of LeGO-LOAM in order to continuously update and refine the complete map. 
 The previous step is repeated with a section of sequence 7 applied on the currently created map, so the final outcome is a merged map built upon sequences of three different days. 

 In this context we can prove the strength of our approach which, by filtering out movable objects, is able to establish more robust connections to previous days maps and more often, therefore obtaining a cleaner and more accurate final representation. 
 A comparison of the quality of the resulting maps, with and without including movable objects can be observed in figure Fig.~\ref{fig:map_extension}.
 Looking at the unfiltered map more blurry regions and doubled elements can be observed which are on the other hand compensated in our solution.

 To get a quantitative measurement of these experiments we are comparing the composed trajectories of the map extension to the corresponding ground truth data. Here we use ICP to align the individual trajectories to the ground truth one, where we are using the resulting fitness score as validation metric. 
 In this investigations it is more important to have correct transformations in between the respective partial sequences rather than achieving a minimum pose-to-pose distance at every time step from the start like in the previous experiment. Therefore we calculate the remaining sum of squared distances after align the composed trajectories to the ground truth. 
 In this regard, we observe that by filtering out movable objects we are able to decrease the trajectory fitting score from $0.26$ to $0.13$, which represents an improvement of $50\%$ compared to the unfiltered approach.



\vspace{-1mm}
\section{CONCLUSION}
In this work we proposed a robust LiDAR-based re-localization algorithm for autonomous driving tasks in highly dynamic environments. By filtering possible \textit{movable} objects based on a convolutional dual-view architecture, we can achieve a more robust, distinctive and static representation of the current environment, which can be used for further processing tasks such as path planning, map updating or re-localization. We proved that by filtering movable objects, the accuracy of re-localization inside a pre-built map can be increased by an average percentage of $35.1\%$ compared to re-localization using the full point clouds and by $47.9\%$ compared to a state-of-the-art lidar odometry and mapping algorithm. Furthermore we showed the adaptability of our approach by applying it to a multi-day map building task, where the accuracy of the final filtered map after applying our method exceeds the ones using full point clouds by $50\%$.





\bibliographystyle{ieeetr}
\bibliography{bib_itsc_loam}

\end{document}